%% file: main.tex
\documentclass[runningheads]{llncs}

\usepackage{eccv}

\usepackage{eccvabbrv}
\usepackage{graphicx}
\usepackage{booktabs}
\usepackage{amsmath}
\usepackage{amssymb}
\usepackage{dsfont}
\usepackage{multirow}
\usepackage{xcolor}

\usepackage[accsupp]{axessibility}

\usepackage{hyperref}

\usepackage{orcidlink}

\begin{document}

\title{Real Eyes Realize Faster: Gaze Stability and Pupil Novelty for Efficient Egocentric Learning} 

\titlerunning{Physiological Frame Curation}

\author{Ajan Subramanian \and Sumukh Bettadapura \and Rohan Sathish}

\authorrunning{A. Subramanian et al.}

\institute{Kubo Technologies}

\maketitle

\begin{abstract}
Always-on egocentric cameras are increasingly used as demonstrations for embodied robotics, imitation learning, and assistive AR, but the resulting video streams are dominated by redundant and low-quality frames. Under the storage and battery constraints of wearable devices, choosing which frames to keep is as important as how to learn from them. We observe that modern eye-tracking headsets provide a continuous, training-free side channel that decomposes into two complementary axes: gaze fixation captures visual stability (quality), while pupil response captures arousal-linked moments (novelty). We operationalize this insight as a Dual-Criterion Frame Curator that first gates frames by gaze quality and then ranks the survivors by pupil-derived novelty. On the Visual Experience Dataset (VEDB), curated frames at 10\% budget match the classification performance of the full stream, and naive signal fusion consistently destroys both contributions. The benefit is task-dependent: pupil ranking improves activity recognition, while gaze-only selection already dominates for scene recognition, confirming that the two signals serve genuinely different roles. Our method requires no model inference and operates at capture time, offering a path toward efficient, always-on egocentric data curation.
\end{abstract}


\input{sections/introduction}

\input{sections/related_work}

\input{sections/decomposition}

\input{sections/method}

\input{sections/setup}

\input{sections/results_mechanistic}

\input{sections/results_efficiency}

\input{sections/results_task_physio}

\input{sections/discussion}

\input{sections/conclusion}

\bibliographystyle{splncs04}

\input{main.bbl}
\newpage
\setcounter{section}{0}
\renewcommand{\thesection}{\Alph{section}}
\part*{Supplementary Material}

\section{Qualitative Examples}
\label{sec:supp_qualitative}

\begin{figure}[h]
    \centering
    \includegraphics[width=0.9\linewidth]{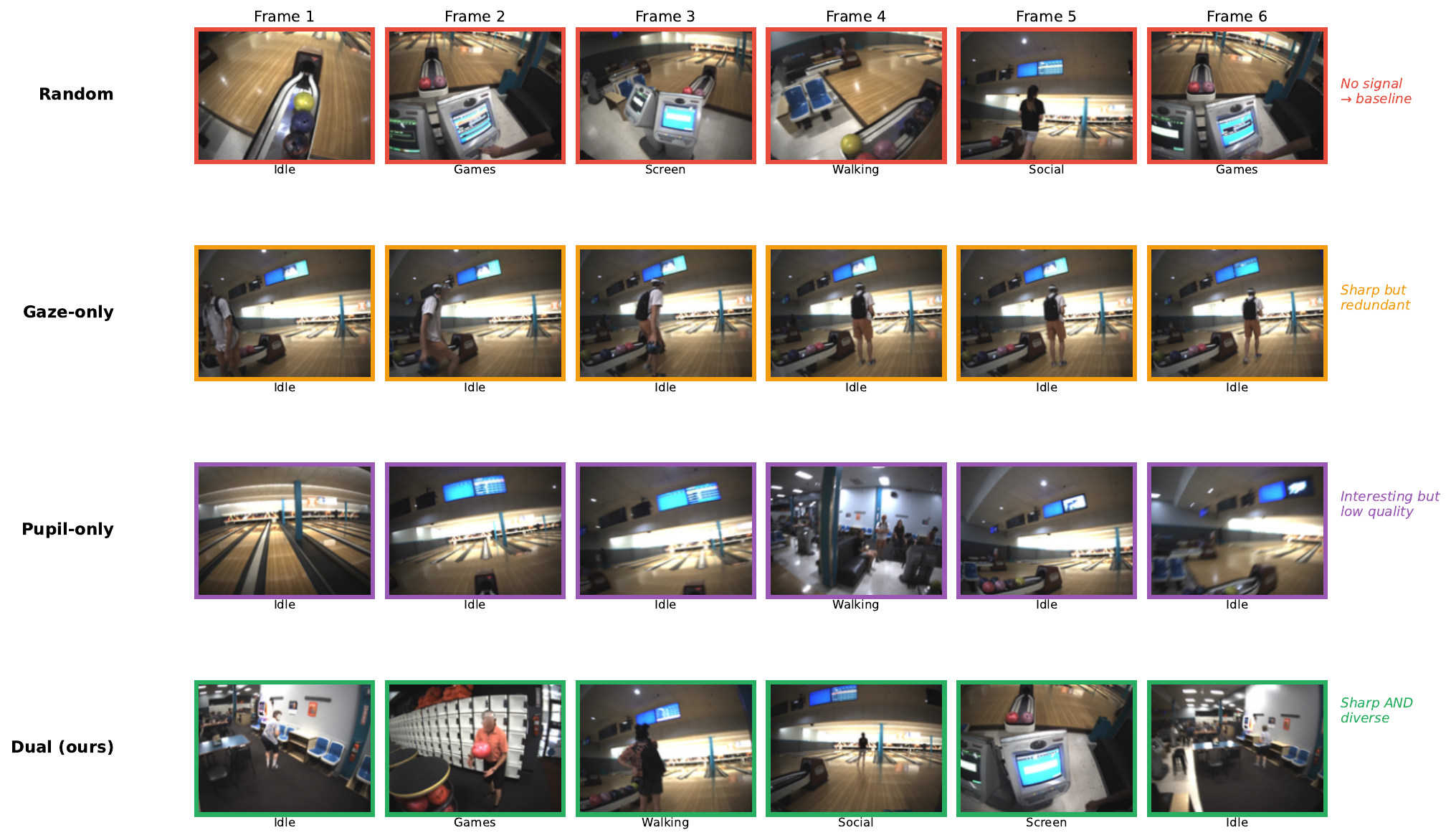}
    \caption{%
    \textbf{Qualitative Comparison.}
    Top: dual-criterion (high gaze, high pupil).
    Bottom: random baseline, including blur and low-information content.
    }
    \label{fig:qualitative}
\end{figure}

Figure~\ref{fig:qualitative} shows representative frames selected by the dual-criterion curator compared to random sampling. The dual-criterion method consistently selects frames with high visual quality (sharp, well-framed) that also capture informationally rich moments such as activity transitions and novel objects entering the scene.

\section{Fine-Grained Classification}
\label{sec:supp_finegrained}

We tested whether physiological curation could help distinguish visually similar activities. Table~\ref{tab:finegrained} reports results for two fine-grained binary tasks evaluated under 5-fold session-level cross-validation.

\begin{table}[h]
\centering
\footnotesize
\caption{Fine-grained classification. No benefit at representation floor (phone vs.\ computer) or for trivially separable categories.}
\label{tab:finegrained}
\begin{tabular}{lcccc}
\toprule
\textbf{Task} & \textbf{Sessions} & \textbf{Best $F1$} & \textbf{Chance} & \textbf{Verdict} \\
\midrule
Phone vs.\ Computer & 31 & 0.507 & 0.50 & At chance \\
Cog.\ vs.\ Phys.\ Games & 15 & 0.912 & 0.50 & Visually separable \\
\bottomrule
\end{tabular}
\end{table}

\paragraph{Phone vs.\ Computer.} Both activities involve staring at a screen, and frozen DINOv2 features cannot separate them regardless of frame selection ($F1 = 0.507$, effectively chance). This represents a representation-level floor that no curation strategy can overcome.

\paragraph{Cognitive vs.\ Physical Games.} This task is solvable ($F1 = 0.912$) but trivially so: sudoku at a desk versus bowling in an alley are visually obvious, and pupil adds nothing because the class-discriminative signal is already in the pixels.

These results define the method's scope: physiological curation is most valuable for coarse classification where activities produce visually distinct but temporally mixed frames. Fine-grained tasks such as object recognition or hand state classification may benefit from different curation criteria (e.g., hand visibility gating).

\section{Classifier Comparison}
\label{sec:supp_classifier}

To verify that our findings are not artifacts of the linear probe, we re-ran all six main-paper strategies using the same 10 seeds and 6 budgets as the main paper, with three classifier families: $L_2$-regularized logistic regression (Linear), a two-layer MLP (768$\rightarrow$256$\rightarrow$classes, ReLU, dropout 0.3), and histogram-based gradient-boosted trees (GBT). The Linear column matches the main paper tables exactly; MLP and GBT columns show how rankings change with classifier choice.

\begin{table}[h]
\centering
\footnotesize
\caption{AULC (mean macro $F1$ across 6 budgets, 10 seeds) per strategy and classifier for \textbf{activity recognition}. Naive fusion ranks near the bottom across all three classifiers.}
\label{tab:clf_activity}
\begin{tabular}{lccc}
\toprule
\textbf{Strategy} & \textbf{Linear} & \textbf{MLP} & \textbf{GBT} \\
\midrule
Dual (gaze75 + pupil) & \textbf{0.223} & 0.220 & 0.133 \\
Gaze-only & 0.218 & 0.225 & 0.135 \\
Pupil-abs & 0.215 & 0.208 & 0.124 \\
Gate + random & 0.205 & \textbf{0.238} & 0.144 \\
Naive fusion & 0.198 & 0.219 & 0.132 \\
Random & 0.197 & 0.224 & \textbf{0.145} \\
\bottomrule
\end{tabular}
\end{table}

\begin{table}[h]
\centering
\footnotesize
\caption{AULC (mean macro $F1$ across 6 budgets, 10 seeds) per strategy and classifier for \textbf{scene recognition}. Gaze-only leads under Linear and MLP; rankings are more variable under GBT.}
\label{tab:clf_scene}
\begin{tabular}{lccc}
\toprule
\textbf{Strategy} & \textbf{Linear} & \textbf{MLP} & \textbf{GBT} \\
\midrule
Gaze-only & \textbf{0.280} & \textbf{0.294} & 0.245 \\
Random & 0.265 & 0.282 & \textbf{0.276} \\
Gate + random & 0.262 & 0.280 & 0.270 \\
Naive fusion & 0.256 & 0.266 & 0.233 \\
Pupil-abs & 0.255 & 0.272 & 0.230 \\
Dual (gaze75 + pupil) & 0.253 & 0.265 & 0.240 \\
\bottomrule
\end{tabular}
\end{table}

\paragraph{Rank correlations.} We measure rank agreement across classifier families using Kendall's $\tau$ over the six strategy AULC values. For activity recognition: Linear vs.\ MLP $\tau = -0.07$, Linear vs.\ GBT $\tau = -0.33$, MLP vs.\ GBT $\tau = 0.73$. For scene recognition: Linear vs.\ MLP $\tau = 0.87$, Linear vs.\ GBT $\tau = 0.47$, MLP vs.\ GBT $\tau = 0.33$. For activity recognition, detailed strategy rankings are highly classifier-dependent, which is expected when the probe evaluates frozen representations rather than learned features. For scene recognition, Linear and MLP show strong agreement ($\tau = 0.87$). The two main claims are robust: (1) for activity recognition, naive fusion ranks 5th out of 6 under all three classifiers; (2) for scene recognition, gaze-only ranks first under both Linear and MLP.

\section{Balanced Class Weight Robustness}
\label{sec:supp_balanced}

Because activity and scene classes are imbalanced, we verified that strategy rankings are preserved when using balanced class weights (inverse-frequency weighting in the logistic regression loss). Table~\ref{tab:balanced} compares AULC under default and balanced settings.

\begin{table}[h]
\centering
\footnotesize
\caption{AULC under default vs.\ balanced class weights (linear probe). Strategy rankings are preserved. $^\ddagger$3 seeds.}
\label{tab:balanced}
\begin{tabular}{lcccc}
\toprule
 & \multicolumn{2}{c}{\textbf{Activity}} & \multicolumn{2}{c}{\textbf{Scene}} \\
\cmidrule(lr){2-3} \cmidrule(lr){4-5}
\textbf{Strategy} & \textbf{Default} & \textbf{Balanced} & \textbf{Default} & \textbf{Balanced} \\
\midrule
Dual (gaze75 + pupil) & \textbf{0.223} & 0.219 & 0.253 & 0.247 \\
Gaze-only & 0.218 & 0.214 & \textbf{0.280} & \textbf{0.269} \\
Pupil-abs & 0.215 & \textbf{0.222} & 0.255 & 0.249 \\
Gate + random & 0.205 & 0.197 & 0.262 & 0.256 \\
Naive fusion$^\ddagger$ & 0.198 & 0.205 & 0.256 & 0.254 \\
Random & 0.197 & 0.204 & 0.265 & 0.258 \\
\bottomrule
\end{tabular}
\end{table}

For scene recognition, the ranking is perfectly preserved (Kendall's $\tau = 1.0$, $p = 0.017$). For activity recognition, the top three strategies swap order slightly ($\tau = 0.4$), but the key structural finding holds: physiologically curated strategies (dual, gaze, pupil) consistently outperform random sampling under both weighting schemes. The absolute $F1$ values change by less than 0.01 on average.

\section{Global vs.\ Stratified Selection}
\label{sec:supp_stratified}

All main paper results use \emph{global} frame selection: ranking frames across all classes without label access, reflecting the training-free deployment scenario. As an oracle comparison, we also tested \emph{stratified} selection, which samples the top-$k$ frames within each class proportionally.

\begin{table}[h]
\centering
\footnotesize
\caption{Global vs.\ stratified at 10\% budget. Stratified helps activity; mixed for scene. $^\ddagger$3 seeds.}
\label{tab:stratified}
\begin{tabular}{llccc}
\toprule
\textbf{Task} & \textbf{Strategy} & \textbf{Global $F1$} & \textbf{Stratified $F1$} & \textbf{$\Delta$} \\
\midrule
\multirow{5}{*}{Activity} 
& Dual (gaze75 + pupil) & 0.228 & 0.248 & $+0.020$ \\
& Gaze-only & 0.213 & 0.224 & $+0.011$ \\
& Random & 0.184 & 0.193 & $+0.008$ \\
& Pupil-abs & 0.161 & 0.188 & $+0.027$ \\
& Naive fusion$^\ddagger$ & 0.164 & 0.198 & $+0.034$ \\
\midrule
\multirow{5}{*}{Scene} 
& Dual (gaze75 + pupil) & 0.270 & 0.268 & $-0.001$ \\
& Gaze-only & 0.321 & 0.282 & $-0.039$ \\
& Random & 0.270 & 0.271 & $+0.002$ \\
& Pupil-abs & 0.248 & 0.242 & $-0.006$ \\
& Naive fusion$^\ddagger$ & 0.269 & 0.269 & $+0.000$ \\
\bottomrule
\end{tabular}
\end{table}

\paragraph{Interpretation.} For activity recognition, stratified selection consistently helps: the dual curator gains $+0.020$ $F1$, and even pupil-abs (which suffers at 10\% budget under global selection) recovers partially under stratification ($+0.027$). Activity classes cluster unevenly across sessions, so balanced per-class sampling prevents minority-class starvation. For scene recognition, the effect is mixed: gaze-only loses substantially under stratification ($-0.039$), while other strategies show negligible differences, suggesting scene identity is more uniformly distributed across the dataset.

\paragraph{Practical Implication.} The main paper results represent a conservative estimate for activity tasks; practitioners with label access can achieve additional gains via stratified curation. For scene recognition, global (training-free) selection remains preferable for gaze-only.

\section{Gate Threshold Sensitivity}
\label{sec:supp_gate}

The main paper uses a 75\% gaze gate (retaining top 75\% of frames by gaze confidence). Table~\ref{tab:gate_full} shows performance across all tested gate thresholds.

\begin{table}[h]
\centering
\footnotesize
\caption{Gate threshold ablation (activity, 10 seeds, 6 budgets). 75\% achieves best F1 at 10\% budget.}
\label{tab:gate_full}
\begin{tabular}{lccccccc}
\toprule
\textbf{Gate \%} & \textbf{5\%} & \textbf{10\%} & \textbf{25\%} & \textbf{50\%} & \textbf{75\%} & \textbf{100\%} & \textbf{Mean} \\
\midrule
25\% & 0.196 & 0.176 & 0.229 & 0.229 & \textbf{0.229} & \textbf{0.229} & 0.214 \\
50\% & 0.132 & 0.225 & 0.237 & 0.205 & 0.205 & 0.205 & 0.201 \\
75\% & \textbf{0.205} & \textbf{0.228} & 0.220 & 0.225 & 0.228 & 0.228 & \textbf{0.223} \\
90\% & 0.167 & 0.149 & 0.246 & 0.232 & 0.185 & 0.216 & 0.199 \\
100\% (no gate) & 0.173 & 0.161 & \textbf{0.257} & \textbf{0.259} & 0.218 & 0.224 & 0.215 \\
\bottomrule
\end{tabular}
\end{table}

\paragraph{Key Observations:}
\begin{itemize}
    \item At low budgets (5--10\%), the 75\% gate is clearly optimal, filtering out noisy frames that would otherwise dominate small samples.
    \item At moderate budgets (25--50\%), relaxing the gate (90\% or no gate) performs better, as there are enough frames to dilute noise.
    \item At high budgets (75--100\%), the strict 25\% gate leads because it retains only the highest-quality frames, which dominate when the pool is large.
    \item The 75\% threshold achieves the best mean performance across all six budgets, confirming it as a robust default.
\end{itemize}

\section{Full Strategy Landscape}
\label{sec:supp_landscape}

Tables~\ref{tab:landscape_act} and~\ref{tab:landscape_scene} report the per-budget macro $F1$ and AULC for all selection strategies evaluated in this work. All rows use 10 seeds except naive fusion ($^\ddagger$), which uses 3 seeds from an earlier experiment; its ranking is deterministic (zero variance across seeds).

\begin{table}[h]
\centering
\footnotesize
\setlength{\tabcolsep}{4pt}
\caption{Per-budget macro $F1$ and AULC for \textbf{activity recognition}, ranked by AULC. $^\ddagger$3 seeds.}
\label{tab:landscape_act}
\begin{tabular}{lcccccc|c}
\toprule
\textbf{Strategy} & \textbf{5\%} & \textbf{10\%} & \textbf{25\%} & \textbf{50\%} & \textbf{75\%} & \textbf{100\%} & \textbf{AULC} \\
\midrule
Dual (75\% + pupil) & 0.205 & \textbf{0.228} & 0.220 & 0.225 & 0.228 & 0.228 & \textbf{0.223} \\
Gaze-only & \textbf{0.207} & 0.213 & 0.230 & 0.204 & 0.227 & 0.225 & 0.218 \\
Pupil-abs & 0.173 & 0.161 & \textbf{0.257} & \textbf{0.259} & 0.218 & 0.224 & 0.215 \\
Gate + random (75\%) & 0.179 & 0.188 & 0.190 & 0.215 & \textbf{0.229} & \textbf{0.229} & 0.205 \\
Dual (50\% + pupil) & 0.132 & 0.225 & 0.237 & 0.205 & 0.205 & 0.205 & 0.201 \\
Naive fusion$^\ddagger$ & 0.177 & 0.164 & 0.190 & 0.208 & 0.223 & 0.225 & 0.198 \\
Random & 0.177 & 0.184 & 0.182 & 0.201 & 0.215 & 0.224 & 0.197 \\
Gate + random (50\%) & 0.177 & 0.177 & 0.215 & 0.204 & 0.204 & 0.204 & 0.197 \\
\bottomrule
\end{tabular}
\end{table}

\begin{table}[h]
\centering
\footnotesize
\setlength{\tabcolsep}{4pt}
\caption{Per-budget macro $F1$ and AULC for \textbf{scene recognition}, ranked by AULC. $^\ddagger$3 seeds.}
\label{tab:landscape_scene}
\begin{tabular}{lcccccc|c}
\toprule
\textbf{Strategy} & \textbf{5\%} & \textbf{10\%} & \textbf{25\%} & \textbf{50\%} & \textbf{75\%} & \textbf{100\%} & \textbf{AULC} \\
\midrule
Gaze-only & \textbf{0.342} & \textbf{0.321} & 0.256 & 0.249 & 0.251 & \textbf{0.260} & \textbf{0.280} \\
Random & 0.294 & 0.270 & 0.257 & \textbf{0.256} & 0.256 & 0.259 & 0.265 \\
Gate + random (75\%) & 0.283 & 0.274 & 0.257 & 0.256 & 0.251 & 0.251 & 0.262 \\
Gate + random (50\%) & 0.289 & 0.271 & 0.252 & 0.249 & 0.249 & 0.249 & 0.260 \\
Dual (50\% + pupil) & 0.259 & 0.277 & 0.271 & 0.249 & 0.249 & 0.249 & 0.259 \\
Naive fusion$^\ddagger$ & 0.261 & 0.269 & 0.238 & 0.252 & \textbf{0.259} & 0.260 & 0.256 \\
Pupil-abs & 0.253 & 0.248 & 0.265 & 0.248 & 0.255 & 0.259 & 0.255 \\
Dual (75\% + pupil) & 0.217 & 0.270 & \textbf{0.280} & 0.253 & 0.250 & 0.250 & 0.253 \\
\bottomrule
\end{tabular}
\end{table}

\paragraph{Key findings.} For activity recognition, the dual curator achieves the highest AULC and leads at 10\% budget, the operating point most relevant for on-device curation. Gaze-only edges ahead at 5\%, while pupil-abs dominates at moderate budgets (25--50\%) where novelty ranking can draw from a larger frame pool but drops at tight budgets where it lacks the quality filter. For scene recognition, gaze-only leads at tight budgets and overall AULC, while random sampling is competitive at moderate-to-high budgets, consistent with the task-dependent analysis in the main paper. All strategies converge at 100\% budget, confirming that differences are driven by frame selection rather than classifier variance.

\section{Dataset Statistics}
\label{sec:supp_dataset}

\begin{table}[h]
\centering
\footnotesize
\caption{VEDB dataset statistics after quality filtering.}
\label{tab:dataset}
\begin{tabular}{lc}
\toprule
\textbf{Statistic} & \textbf{Value} \\
\midrule
Total sessions & 717 \\
Sessions with eye tracking & 136 \\
Sessions with synchronized gaze + pupil & 119 \\
Unique observers & 56 \\
Age range & 7--46 years \\
Total frames (1 fps) & 154,819 \\
Activity classes & 12 \\
Scene classes & 16 \\
Mean session duration & 19.2 min \\
\bottomrule
\end{tabular}
\end{table}

\paragraph{Activity Classes:} Walking, cooking, driving, screen use, reading, eating, socializing, shopping, exercising, playing games, grooming, chores.

\paragraph{Scene Classes:} Kitchen, office, street, store, restaurant, gym, park, bedroom, living room, bathroom, garage, yard, car interior, public transit, classroom, other.

\section{Implementation Details}
\label{sec:supp_implementation}

\paragraph{Pupil Normalization.}
Raw pupil diameter is normalized per-session using:
\begin{enumerate}
    \item 10-second rolling median for local baseline subtraction (robust to blinks)
    \item MAD-based z-scoring: $p_z = (p - \text{median}) / \text{MAD}$
\end{enumerate}

\paragraph{Gaze Confidence.}
Gaze confidence values (0--1) are provided directly by the Pupil Labs API, computed from the 3D eye model fit quality.

\paragraph{Feature Extraction.}
DINOv2 ViT-B/14 features are extracted using the \texttt{torch.hub} pretrained model with default preprocessing (224$\times$224 center crop, ImageNet normalization).

\paragraph{Reproducibility.}
Raw VEDB data is available at \url{https://osf.io/4mpz5/}.

\end{document}

%% file: sections/introduction.tex
\section{Introduction}
\label{sec:intro}

Always-on egocentric cameras promise a complete digital memory and are increasingly used as a scalable source of demonstrations for embodied robotics and assistive AR~\cite{hoque2025egodex, kareer2025egomimic}. However, practical streams are dominated by redundancy and uninformative content like blinks, motion blur, and static scenes. With massive datasets now containing thousands of hours of video~\cite{grauman2022ego4d, damen2022epic100}, the bottleneck is deciding which frames to keep under strict storage and battery budgets~\cite{zha2023datacentric}. Fortunately, modern AR headsets and research glasses increasingly integrate eye tracking~\cite{apple_vision_pro_specs, ms_hololens_eye_tracking, engel2023aria}, transforming curation into a multimodal sensing problem. By treating the eye as a side channel during capture, we can select which frames to keep without running any vision model, reducing the data that must be stored, transmitted, and labeled.

Current strategies fail this efficiency paradox. Random sampling wastes budget on blur, while diversity-driven coresets~\cite{sener2018coreset, gygli2015submod} require computationally prohibitive feature extraction. We therefore focus on physiological signals available at capture time. Gaze is a strong candidate for filtering image quality, as steady fixation correlates with visual stability~\cite{fathi2012gaze}. However, stability creates a new failure mode. High gaze confidence often targets low-information segments (e.g., reading), collapsing the stream into redundancy. Ranking frames solely by gaze yields crisp images that are semantically identical.

To address this, we propose a Quality-Novelty decomposition. Gaze provides the quality axis, but we need a second signal for novelty. Pupil response reflects arousal and cognitive engagement~\cite{kahneman1966pupil, preuschoff2011pupil}, with dilation linked to attention shifts and informative moments~\cite{joshi2016relationships}. We hypothesize that these two signals serve complementary roles, quality filtering versus novelty ranking, and that their combination requires careful design. To test this, we introduce a \textbf{Dual-Criterion Frame Curator} that applies a Gaze Quality Gate followed by a Pupil Novelty Ranker, and compare it against single-signal baselines and naive fusion. On the Visual Experience Dataset (VEDB)~\cite{greene2024visual}, the dual curator selects 10\% of frames that match full-stream classification performance for activity recognition, with no model computation in the selection loop.

Our contributions are as follows:
\begin{enumerate}
\item We formalize curation as a Quality-Novelty decomposition, positioning gaze as a stability proxy and pupil dynamics as a novelty proxy.
\item We propose a Dual-Criterion Frame Curator that gates by gaze quality and ranks by pupil novelty to select high-value frames.
\item On VEDB, we show that 10\% of physiologically curated frames matches full-stream performance for activity recognition, that naive fusion of the two signals destroys both, and that the benefit is task-dependent: pupil ranking helps for activities but not scenes.
\end{enumerate}

\begin{figure}[t]
    \centering
    \includegraphics[width=0.95\linewidth]{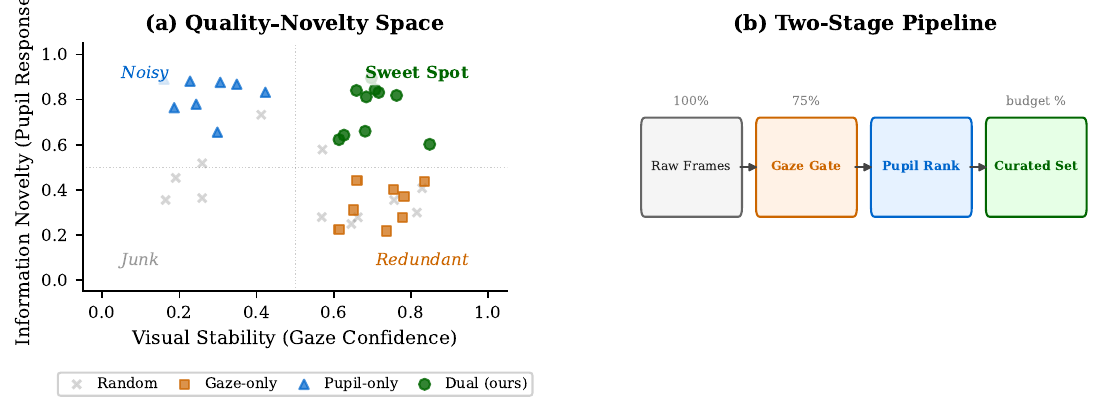}
    \caption{%
    \textbf{Quality--Novelty Decomposition.}
    \emph{(a)} Gaze confidence (x) captures stability; pupil response (y) captures novelty.
    Random includes junk; gaze-only yields clean but redundant frames; dual targets high stability \emph{and} novelty.
    \emph{(b)} Two-stage pipeline: gaze gate (top 75\%) $\rightarrow$ pupil ranking within budget.
    }
    \label{fig:quadrant}
\end{figure}

%% file: sections/related_work.tex
\section{Related Work}
\label{sec:related}

\textbf{Efficient video sampling and curation.}
Large-scale egocentric datasets such as Ego4D~\cite{grauman2022ego4d} and EPIC-Kitchens~\cite{damen2022epic100} contain thousands of hours of video, yet most frames are redundant or uninformative. Selecting a compact, high-quality subset is a longstanding problem studied under several names. Video summarization methods learn to pick keyframes or keyshots that capture the narrative of a clip~\cite{gygli2014summe,song2015tvsum,zhang2016lstm,gong2014seqdpp}, but they optimize for human viewing rather than downstream classification. Coreset and data pruning approaches select training subsets that preserve model accuracy~\cite{toneva2019empirical,coleman2020selection}, and at web scale, filtering pipelines have shown that curated data can outperform larger unfiltered pools~\cite{gadre2023datacomp}. Active learning reduces labeling cost by querying the most informative examples~\cite{settles2009active,sener2018coreset}, yet both coreset selection and active learning typically require computing embeddings or model scores over the entire data pool. Recent keyframe selection methods for long-video understanding similarly rely on learned relevance estimates~\cite{alfasly2023fastpicker,zhu2026focus}. The common bottleneck is that selection happens after feature extraction, making these approaches unsuitable as a capture-time curator on resource-constrained wearable devices where the decision of what to keep must precede any model computation.

\textbf{Gaze as a supervision and attention signal.}
Gaze has a long history as a cue for egocentric recognition. Early work showed that fixation patterns help identify objects and actions from first-person video~\cite{fathi2012gaze,li2015delving}. Subsequent methods integrate gaze into spatial attention mechanisms for activity recognition~\cite{min2021gaze} or predict gaze from video to guide downstream tasks~\cite{huang2018gaze,lai2024eye}. However, these approaches use gaze primarily as an input feature or supervisory signal during training, and recent studies question whether straightforward gaze fusion reliably improves egocentric action recognition, reporting failure modes when network attention poorly overlaps with fixation~\cite{zhang2022gazeinform}. Our use of gaze differs: rather than feeding gaze into a model, we use it to \emph{construct a better training set} by filtering out low-quality frames before any model is trained.

\textbf{Pupilometry as an arousal and novelty proxy.}
Pupil diameter tracks task-evoked cognitive effort and arousal~\cite{beatty1982task,kahneman1966pupil,mathot2018pupillometry}, with dilation linked to surprise and uncertainty~\cite{preuschoff2011pupil} through the locus coeruleus--norepinephrine system~\cite{astonjones2005integrative,joshi2016relationships,reimer2016pupil}. Applied studies use pupil responses to index workload and fatigue in real workflows~\cite{kremer2024workload}, and memory research shows that pupil dilation during encoding predicts later recall, consistent with a novelty signal~\cite{kafkas2024eyes}. However, prior vision work rarely treats pupil as a frame selection signal for constructing training sets, and no existing method combines gaze and pupil in a way that respects their different roles. The gap is not whether either signal is useful in isolation, but that their \emph{complementarity} for budgeted, capture-time curation remains unexploited.

%% file: sections/decomposition.tex
\section{From Physiological Signals to Frame Scores}
\label{sec:frame_scores}

Our curator operates at the level of individual video frames. For each frame timestamp $t$, we compute two scalars using only eye tracker outputs available at recording time: a gaze-based \textbf{quality score} $g(t)$ and a pupil-based \textbf{novelty score} $p(t)$.

\paragraph{Frame-level alignment.}
Eye trackers sample at rates much higher than the video frame rate. We sample video frames at $1$\,fps and align the eye tracker stream by aggregating all measurements within a $\pm 50$\,ms window centered on each frame timestamp $t$. The scores defined below are computed from these aggregated samples.

\paragraph{Gaze Quality Score $g(t)$.}
We define quality as the product of fixation state and tracking confidence:
\begin{equation}
    g(t) = f(t) \cdot c(t)
    \label{eq:gaze}
\end{equation}
where $c(t) \in [0,1]$ is the mean detection confidence reported by the eye tracker, and $f(t) \in [0,1]$ is a soft fixation indicator computed in two steps. First, we measure the fraction of gaze samples in the window whose velocity falls below 0.5 normalized units/s. Second, we scale this fraction by $\min(1,\, d / 150\text{\,ms})$, where $d$ is the total duration of the low-velocity interval. This linear ramp gives proportional credit to shorter fixations and prevents brief glances (e.g., 50\,ms) from receiving the same score as sustained fixation. When $g(t)$ is high, the wearer's gaze is steady and the tracker is confident, indicating that the frame is likely sharp and well-observed.

\paragraph{Pupil Novelty Score $p(t)$.}
Raw pupil diameter is dominated by the pupillary light reflex. Following standard preprocessing guidelines~\cite{kret2019preprocessing}, we remove this luminance component and normalize the residual in three steps:
\begin{enumerate}
    \item \textbf{Luminance correction.} Per recording session, we fit a polynomial regression of pupil diameter on scene brightness and subtract the predicted values, isolating the cognitively-driven component.
    \item \textbf{Local baseline removal.} We subtract a 10-second rolling median to remove slow drift (e.g., fatigue, arousal trends), capturing deviations relative to the recent past.
    \item \textbf{Robust normalization.} We z-score with median and MAD (robust to heavy-tailed pupil outliers), yielding a zero-centered, unit-scale signal where $|p(t)| > 2$ marks salient deviations.
\end{enumerate}
The result is a cleaned pupil signal $p(t)$. We use the \emph{absolute value} $|p(t)|$ as our novelty measure: both dilation (arousal, surprise) and constriction (cognitive effort, bright transitions) mark moments where the visual environment changed or the wearer's attention shifted~\cite{kahneman1966pupil, preuschoff2011pupil, joshi2016relationships}. Taking the magnitude captures both directions.

\paragraph{Temporal alignment of pupil responses.}
The pupil response has biological latency: dilation typically peaks 300--1500\,ms after a triggering stimulus~\cite{hoeks1993pupillary}. To align the novelty score with the visual content that evoked it, we define a \emph{delayed} variant that computes $p(t)$ from samples in a forward-shifted window $[t + 300\text{\,ms},\, t + 1500\text{\,ms}]$. We also evaluate a \emph{no-delay} variant (centered on $t$) in our experiments.

%% file: sections/method.tex
\section{Dual-Criterion Frame Curator}
\label{sec:method}

Given a frame stream scored by quality $g(t)$ and novelty $|p(t)|$ (Section~\ref{sec:frame_scores}), we want to select a subset of $b\%$ of frames that are both visually usable and informatively diverse. With two per-frame scores, there are three natural designs: use a single score, fuse both into one scalar and rank (as in diversity-based selection~\cite{kulesza2012dpp}), or compose them sequentially. We consider each in turn.

\paragraph{Single-signal baselines.}
Selecting frames with the highest $g(t)$ removes junk (blinks, blur, tracking failures) but over-selects static, well-fixated content. Extended fixations on familiar objects yield both high confidence and high stability, so the top-ranked frames tend to be semantically repetitive. Selecting frames with the largest $|p(t)|$ captures moments of arousal and engagement, but without quality control the selection can include blurry frames, saccadic smear, or tracking dropouts that happen to coincide with pupil events. Each signal addresses one side of the problem but not both.

\paragraph{Naive fusion.}
A natural alternative is to combine both scores into a single ranking scalar, for example a weighted sum $w_g \cdot g(t) + w_p \cdot |p(t)|$, and select the top $b\%$. We evaluate this as a baseline with equal weights.

\paragraph{Sequential composition: the Dual-Criterion Curator.}
Instead of fusing scores, we compose them sequentially, analogous to the cascade principle in detection~\cite{viola2004robust}: quality acts as a hard constraint (gate) and novelty acts as a budget allocation rule (rank). The pipeline has two parameters: a \emph{gate fraction} $k$ that controls how aggressively we filter for quality, and a \emph{data budget} $b$ that sets the final number of selected frames. Since $b \leq k$, the novelty ranker always selects from a pool larger than the final budget.

\textbf{Stage~1: Gaze Quality Gate.}
From the full frame set $\mathcal{F}$, retain the top $k\%$ of frames ranked by $g(t)$:
\begin{equation}
    \mathcal{F}_{\text{gated}} = \{ t \in \mathcal{F} : g(t) \geq \tau_k \}
    \label{eq:gate}
\end{equation}
where $\tau_k$ is the per-session threshold such that $k\%$ of frames satisfy $g(t) \geq \tau_k$. This removes frames captured during blinks, saccades, and low-confidence tracking. We set $k = 75$ by default (retain the top 75\%, discard the worst 25\%) and ablate this choice in our experiments.

\textbf{Stage~2: Pupil Novelty Ranker.}
Within the gated pool, rank by $|p(t)|$ and select the top $n$ frames:
\begin{equation}
    \mathcal{F}_{\text{selected}} = \text{top-}n\big( \mathcal{F}_{\text{gated}},\, |p(t)| \big), \quad n = b \cdot |\mathcal{F}|
    \label{eq:rank}
\end{equation}
where $b$ is the fraction of the original stream to retain (the data budget) and $|\mathcal{F}|$ is the total number of candidate frames. The critical experimental control is \textbf{Gate+Random}: the same gaze gate followed by uniform random selection within the gated pool. If our method outperforms this control, the pupil ranking adds genuine signal beyond quality filtering alone.

\paragraph{Strategy summary.}
Table~\ref{tab:strategies} defines all strategies compared in our evaluation. They differ only in how they map $(g(t),\, |p(t)|)$ to a selected subset.

\begin{table}[h!]
\centering
\footnotesize
\caption{Frame selection strategies. Gate+Random is the ablation control that isolates the contribution of pupil ranking.}
\label{tab:strategies}
\begin{tabular}{llll}
\toprule
\textbf{Strategy} & \textbf{Stage 1 (Gate)} & \textbf{Stage 2 (Rank)} & \textbf{Role} \\
\midrule
Random & --- & Uniform random & Baseline \\
Gaze-only & --- & Rank by $g(t)$ & Quality baseline \\
Pupil-abs & --- & Rank by $|p(t)|$ & Novelty baseline \\
Combined & --- & $0.5\,g + 0.5\,|p|$ & Naive fusion \\
\textbf{Dual} & Top 75\% by $g(t)$ & Rank by $|p(t)|$ & \textbf{Proposed} \\
Gate+Random & Top 75\% by $g(t)$ & Uniform random & Ablation \\
\bottomrule
\end{tabular}
\end{table}

%% file: sections/setup.tex
\section{Experimental Setup}
\label{sec:setup}

\paragraph{Dataset.}
We evaluate on the Visual Experience Dataset (VEDB)~\cite{greene2024visual}, which contains 717 egocentric video sessions recorded by 56 observers (ages 7--46) wearing Pupil Labs eye trackers in everyday environments. After filtering for data quality and label availability, we retain 136 sessions (119 with synchronized eye tracking), producing 154,819 frames sampled at 1\,fps. Sessions span 12 activity categories and 16 scene types across indoor, outdoor, active, and sedentary conditions.

\paragraph{Feature Extraction.}
We extract DINOv2 ViT-B/14~\cite{oquab2023dinov2,dosovitskiy2021vit} CLS tokens (768 dimensions) for every frame using the frozen pretrained backbone. No fine-tuning is performed. This is a deliberate choice: our question is whether \emph{frame selection} matters, not whether a particular backbone can be optimized.

\paragraph{Downstream Tasks.}
We evaluate on two classification tasks assigned at the session level: activity recognition (12 classes: walking, cooking, driving, screen use, reading, eating, socializing, shopping, exercising, playing games, grooming, chores) and scene recognition (16 classes: kitchen, office, street, store, etc.).

\paragraph{Frame Selection.}
All curation strategies select frames globally across the training set without access to class labels. This reflects the intended deployment scenario where physiological curation occurs before annotation. We also evaluate stratified selection (per-class sampling) as an oracle comparison; results are discussed in Section~\ref{sec:discussion}.

\paragraph{Protocol.}
We evaluate at data budgets of 5\%, 10\%, 25\%, 50\%, 75\%, and 100\% of training frames. The classifier is $L_2$-regularized logistic regression~\cite{alain2017understanding}; nonlinear models (MLP, SVM, boosted trees) yield comparable results (supplementary). Evaluation uses macro F1 with session-level train/test splits to prevent frame leakage. All stochastic strategies are evaluated over 10 random seeds. Statistical comparisons use one-sample $t$-tests at each budget (df = 9 for stochastic vs.\ deterministic comparisons) and bootstrap confidence intervals (1,000 resamples) on the Area Under the Learning Curve (AULC), the integral of F1 over budget fractions. In total, our evaluation comprises over 5,200 runs across the 6 core strategies and their gate, classifier, and sampling variants (supplementary).

%% file: sections/results_mechanistic.tex
\section{Results: Linking Physiological Signals to Feature-Space Change}
\label{sec:mechanistic}

\begin{figure}[t]
    \centering
    \includegraphics[width=\linewidth]{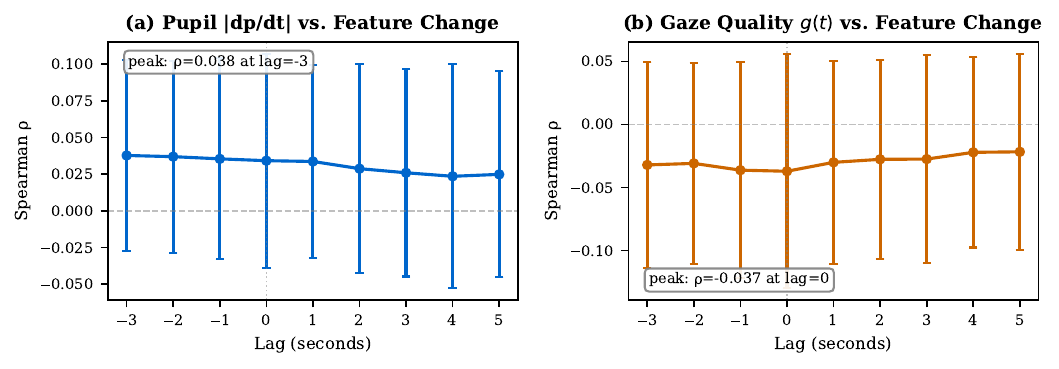}
    \caption{%
    \textbf{Correlation between physiological signals and DINOv2 feature change.}
    \emph{(a)} Pupil derivative $|dp/dt|$ is positively correlated with feature change at all lags (mean $\rho = +0.038$).
    \emph{(b)} Gaze quality $g(t)$ is negatively correlated ($\rho = -0.037$), confirming it tracks stability.
    Error bars: $\pm 1$ s.d.\ across sessions.
    }
    \label{fig:mechanistic}
\end{figure}

Our curation strategy assumes that gaze quality $g(t)$ selects stable frames and pupil novelty $|p(t)|$ selects changing ones. Before evaluating downstream task performance, we verify this directly: do these signals actually correlate with how much DINOv2 features change between consecutive frames?

We define frame-level feature change as
\begin{equation}
    \Delta(t) = 1 - \cos\!\big(\phi(t),\, \phi(t{-}1)\big)
    \label{eq:feature_change}
\end{equation}
where $\phi(t)$ is the DINOv2 CLS embedding at frame $t$. High $\Delta(t)$ means the visual content changed substantially between consecutive frames. Since $\Delta(t)$ is itself a temporal derivative (frame-to-frame change), we correlate it with the absolute pupil derivative $|dp/dt|$ rather than the magnitude $|p(t)|$: the rate of pupil change captures instantaneous arousal fluctuations that co-vary with sudden visual transitions. For gaze, we correlate $\Delta(t)$ with the quality score $g(t)$ directly.

\paragraph{Lag profiles.}
We compute per-session Spearman correlations between the physiological signal at time $t + \ell$ and $\Delta(t)$, where lag $\ell$ ranges from $-3$ to $+5$ seconds (positive $\ell$ means the signal is measured \emph{after} the feature change). Figure~\ref{fig:mechanistic} shows the results averaged across 119 sessions.

The pupil derivative is positively correlated with feature change at every lag tested (mean Spearman $\rho = +0.038$ across sessions at $\ell = -3$\,s, declining to $+0.025$ at $\ell = +5$\,s), indicating a broad temporal association between pupil fluctuations and visual transitions. This broad profile is consistent with sustained arousal responses that span several seconds around salient events. Gaze quality shows the opposite pattern: a negative correlation at all lags (mean $\rho = -0.037$ across sessions at $\ell = 0$), confirming that high $g(t)$ co-occurs with visual stability, not change. Although the individual effect sizes are small ($|\rho| < 0.04$), the directional pattern is consistent across the majority of sessions, which is the relevant test for a per-frame selection signal.

\paragraph{Consequence for curator design.}
The opposing signs confirm that gaze and pupil capture different ends of the feature-change spectrum: gaze selects stable, low-change frames while pupil activity coincides with diverse, high-change ones. Naively averaging two signals that point in opposite directions should be counterproductive, and our experiments confirm this: naive fusion ranks near the bottom of all strategies tested (Sec.~\ref{sec:results_efficiency}). Instead, we apply them sequentially, using gaze to gate for quality first and then pupil to rank the surviving frames by novelty, preserving both contributions without cancellation.

%% file: sections/results_efficiency.tex
\section{Results: Sample Efficiency}
\label{sec:results_efficiency}

\paragraph{Setting expectations.}
At 100\% budget, all non-gated strategies use every available frame and converge to roughly the same macro F1 ($\approx 0.224$ for activity, $\approx 0.25$ for scene). This is expected: when all frames are used, selection is irrelevant. Note that gated strategies (dual, gate+random) at 100\% budget are capped by the gate size (75\% of frames), so they do not reach the same pool. The value of curation lies in the \emph{constrained regime}, specifically budgets of 5--25\%, where on-device storage, battery, and labeling budgets force a choice. Our evaluation focuses on this regime.

\subsection{Main Results}
\label{sec:main_results}

We present learning curves for both tasks below, then focus our detailed statistical analysis on activity recognition, where the dual-criterion curator shows its strongest effect. The scene recognition results, where gaze-only dominates, are analyzed in depth in Section~\ref{sec:task_dependent}.

\begin{figure}[t]
    \centering
    \includegraphics[width=\linewidth]{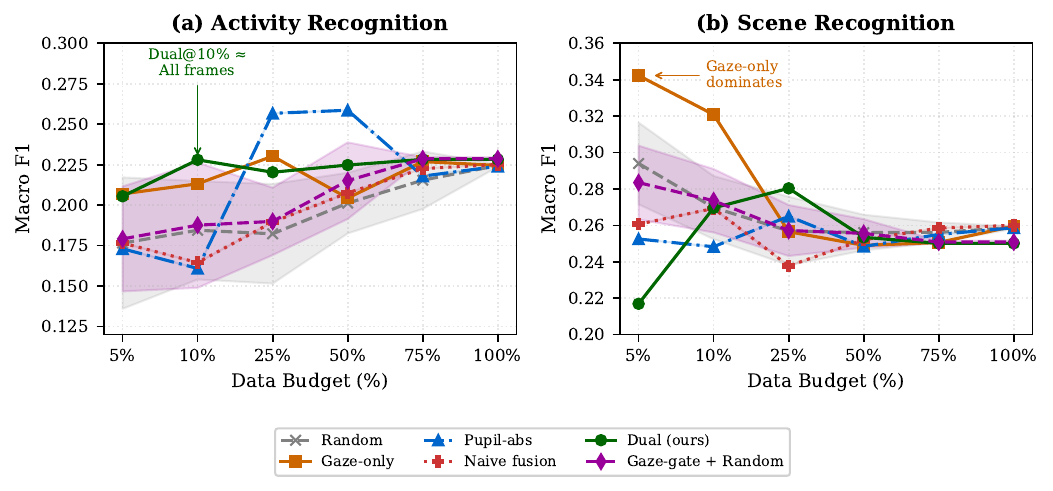}
    \caption{Learning curves: activity (left), scene (right). Dual at 10\% budget matches the performance achieved using all frames. Shaded: $\pm 1$ s.d.\ (10 seeds).}
    \label{fig:learning_curves}
\end{figure}

\paragraph{Activity recognition.}
Figure~\ref{fig:learning_curves} plots macro F1 versus data budget for the core strategies (error bands: $\pm 1$ standard deviation across 10 seeds). The dual-criterion curator at 10\% budget achieves $F1 = 0.228$, exceeding the all-frames baseline ($F1 \approx 0.224$) with one-tenth of the data. Gaze-only achieves $F1 = 0.213$ at the same budget, while the 10-seed random mean yields only $F1 = 0.184$. Table~\ref{tab:aulc} reports the AULC summary with bootstrap confidence intervals and pairwise comparisons. The dual-criterion curator significantly outperforms random ($\Delta\text{AULC} = +0.025$, 95\% CI $[+0.017, +0.033]$, $p < 0.001$). All reported $p$-values survive Bonferroni correction for the four primary comparisons ($\alpha = 0.0125$).

\paragraph{Scene recognition.}
For scene recognition, the picture is different. At 5\% budget, gaze-only achieves $F1 = 0.342$, well above random ($F1 = 0.294$) and dual ($F1 = 0.217$). This advantage persists across budgets: gaze-only has the highest AULC ($0.280$), while adding pupil ranking hurts performance (dual AULC $= 0.253$, below random at $0.265$). This asymmetry is predicted by the quality-novelty decomposition: scene identity is a spatial property captured by stable fixation, not by temporal novelty. We analyze this task dependence further in Section~\ref{sec:task_dependent}.

\begin{table}[h!]
\centering
\footnotesize
\setlength{\tabcolsep}{4pt}
\caption{AULC (mean macro $F1$ across budgets 5\%--100\%) for activity recognition, 10 seeds. $^\dagger$Gate+Rand: gaze gate, random within pool. $^\ddagger$3 seeds; deterministic ranking (zero variance).}
\label{tab:aulc}
\begin{tabular}{lcccc}
\toprule
\textbf{Strategy} & \textbf{AULC} & \textbf{95\% CI} & \textbf{vs.\ Random} & \textbf{$p$} \\
\midrule
\textbf{Dual} & \textbf{0.223} & $[0.223, 0.223]$ & $+0.025$ & $< 0.001$ \\
Gaze-only & 0.218 & $[0.218, 0.218]$ & $+0.020$ & $< 0.001$ \\
Pupil-abs & 0.215 & $[0.215, 0.215]$ & $+0.018$ & $< 0.001$ \\
Gate+Random$^\dagger$ & 0.205 & $[0.196, 0.215]$ & $+0.007$ & $0.210$ \\
Naive fusion$^\ddagger$ & 0.198 & $[0.198, 0.198]$ & $+0.000$ & $1.000$ \\
Random & 0.197 & $[0.189, 0.205]$ & --- & --- \\
\bottomrule
\end{tabular}
\end{table}

\paragraph{Naive fusion.}
Naive fusion (equal-weighted combination of $g(t)$ and $|p(t)|$) achieves an AULC of $0.198$, essentially tied with random ($0.197$) and well below gaze-only ($0.218$) and dual ($0.223$). The failure is sharpest at the critical 10\% budget, where fusion achieves $F1 = 0.164$ compared to $0.213$ for gaze-only and $0.184$ for random. Because the two signals pull in opposing directions (gaze selects stability, pupil selects change), combining them into a single scalar dilutes the gaze contribution without gaining the novelty benefit. This pattern holds across multiple weight configurations (0.3/0.7, 0.5/0.5, 0.7/0.3) and across all three classifier families tested (supplementary), confirming the design choice to compose the signals sequentially rather than fuse them.

\subsection{The Critical Ablation}
\label{sec:ablation}

The most important test compares the dual-criterion curator against Gate+Random: same 75\% gaze gate, but uniform random selection within the gated pool. If pupil ranking outperforms random-within-gate, the novelty signal adds genuine value beyond quality filtering. Table~\ref{tab:ablation} reports the per-budget comparison for activity recognition.

\begin{table}[h!]
\centering
\footnotesize
\setlength{\tabcolsep}{3pt}
\caption{Ablation: dual vs.\ gate+random for activity recognition. Pupil ranking wins at 5--25\% budgets.}
\label{tab:ablation}
\begin{tabular}{lccccc}
\toprule
\textbf{Budget} & \textbf{Dual} & \textbf{Gate+Rand} & \textbf{$\Delta F1$} & \textbf{$p$} & \textbf{Cohen's $d$} \\
\midrule
5\% & 0.205 & $0.179 \pm 0.031$ & $+0.026$ & $0.030$ & 0.86 \\
10\% & 0.228 & $0.188 \pm 0.037$ & $+0.040$ & $0.009$ & 1.10 \\
25\% & 0.220 & $0.190 \pm 0.020$ & $+0.030$ & $0.001$ & 1.54 \\
50\% & 0.225 & $0.215 \pm 0.022$ & $+0.010$ & $0.230$ & 0.43 \\
75\% & 0.228 & $0.229 \pm 0.001$ & $-0.000$ & $0.207$ & --- \\
\bottomrule
\end{tabular}
\end{table}

At budgets of 5--25\%, pupil ranking is significantly better than random-within-gate ($p = 0.030$ at 5\%, $p = 0.009$ at 10\%, $p = 0.001$ at 25\%), with large effect sizes (Cohen's $d > 0.8$). At 75--100\%, the gate passes so many frames that both strategies produce nearly identical subsets, and the effect vanishes. The aggregate AULC comparison confirms: $\Delta\text{AULC} = +0.018$, $p < 0.001$.

\paragraph{Decomposing the improvement.}
The dual-criterion's total advantage over random ($+0.025$ AULC) decomposes into two additive contributions: the gaze gate accounts for $+0.007$ AULC (29\%), while pupil ranking accounts for $+0.018$ AULC (71\%, $p < 0.001$). The pupil component is the larger contributor, confirming that novelty ranking provides genuine value beyond quality filtering.

\subsection{Sensitivity Analysis}
\label{sec:sensitivity}

\paragraph{Regime-dependent crossover.}
\label{sec:crossover}
At 25\% budget, ungated pupil-abs ($F1 = 0.257$) outperforms the dual curator ($F1 = 0.220$). At very low budgets (5--10\%), the gaze gate is essential: without it, pupil ranking selects noisy, blurry frames alongside genuinely novel ones, and pupil-abs achieves only $F1 = 0.161$ at 10\%. At moderate budgets (25--50\%), the gate becomes restrictive, excluding some genuinely informative content. This crossover reveals that the dual-criterion curator is the right tool for the most practically important regime: on-device curation at 5--15\% budget, where storage and battery constraints bite hardest. At larger budgets, the gate threshold should be relaxed.

\paragraph{Gate threshold ablation.}
\label{sec:gate_ablation}
To verify that the 75\% gate is not a lucky threshold, we test gate retention rates of 25\%, 50\%, 75\%, 90\%, and 100\% (no gate $\equiv$ pupil-abs) for activity recognition. Table~\ref{tab:gate} reports the results at three representative budgets.

\begin{table}[h!]
\centering
\footnotesize
\setlength{\tabcolsep}{6pt}
\caption{Gate threshold ablation for activity recognition. 75\% is optimal at 10\% budget.}
\label{tab:gate}
\begin{tabular}{lcccc}
\toprule
\textbf{Gate \%} & \textbf{$b{=}10\%$} & \textbf{$b{=}25\%$} & \textbf{$b{=}50\%$} & \textbf{Mean} \\
\midrule
25\% & 0.176 & 0.229 & 0.229 & 0.211 \\
50\% & 0.225 & 0.237 & 0.205 & 0.222 \\
75\% & \textbf{0.228} & 0.220 & 0.225 & 0.224 \\
90\% & 0.149 & 0.246 & 0.232 & 0.209 \\
100\% (no gate) & 0.161 & \textbf{0.257} & \textbf{0.259} & \textbf{0.225} \\
\bottomrule
\end{tabular}
\end{table}

The 75\% threshold achieves the best F1 at 10\% budget ($F1 = 0.228$), the regime where curation matters most. The no-gate condition achieves the highest overall mean (0.225) by excelling at 25--50\%, but it collapses at 10\% ($F1 = 0.161$). Overly strict gates (25\%) hurt across the board by discarding too many informative frames. This confirms the regime-dependent pattern: the gate's value is concentrated at the lowest budgets.

%% file: sections/results_task_physio.tex
\section{Results: Task-Dependent Physiology}
\label{sec:task_dependent}

\subsection{Activity vs.\ Scene: Pupil Helps One, Hurts the Other}
\label{sec:activity_vs_scene}

\begin{figure}[t]
    \centering
    \includegraphics[width=\linewidth]{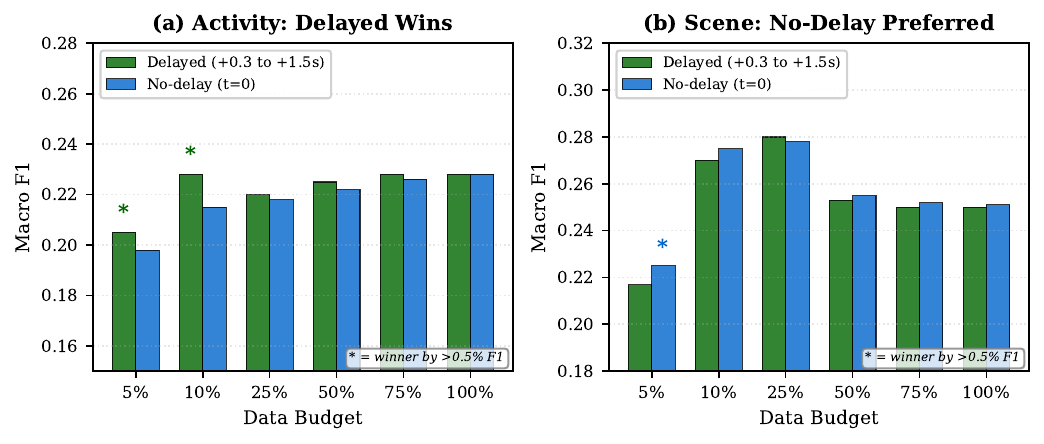}
    \caption{%
    \textbf{Task-Dependent Performance.}
    \emph{(a)} Activity: pupil ranking improves over gaze-only and random.
    \emph{(b)} Scene: gaze-only dominates; pupil adds no benefit.
    }
    \label{fig:task_physio}
\end{figure}

The pupil signal's utility is sharply task-dependent. Table~\ref{tab:task_split} and Figure~\ref{fig:task_physio} summarize the contrast between activity and scene recognition.

\begin{table}[t]
\centering
\footnotesize
\caption{Task-dependent pupil effect. Helps activity ($p < 0.001$); not scene.}
\label{tab:task_split}
\begin{tabular}{lcccc}
\toprule
\textbf{Task} & \textbf{Best Strategy} & \textbf{Dual vs.\ Random} & \textbf{$p$} & \textbf{Pupil Helps?} \\
\midrule
Activity (12-class) & Dual & $+0.025$ AULC & $< 0.001$ & Yes \\
Scene (16-class) & Gaze-only & $-0.012$ AULC & $1.000$ & No \\
\bottomrule
\end{tabular}
\end{table}

For activity recognition, the dual-criterion curator significantly outperforms random ($\Delta\text{AULC} = +0.025$, $p < 0.001$), and the pupil ablation confirms genuine novelty signal ($p < 0.001$ vs.\ gate+random). For scene recognition, the ranking inverts: gaze-only is the strongest strategy (Table~\ref{tab:scene_aulc}), significantly outperforming random ($\Delta\text{AULC} = +0.015$, $p < 0.001$), while adding pupil ranking hurts performance (dual AULC $= 0.253$, below random at $0.265$).

\begin{table}[h!]
\centering
\footnotesize
\setlength{\tabcolsep}{4pt}
\caption{AULC (mean macro $F1$ across budgets 5\%--100\%) for scene recognition, 10 seeds. $^\ddagger$3 seeds; deterministic ranking (zero variance).}
\label{tab:scene_aulc}
\begin{tabular}{lcccc}
\toprule
\textbf{Strategy} & \textbf{AULC} & \textbf{95\% CI} & \textbf{vs.\ Random} & \textbf{$p$} \\
\midrule
\textbf{Gaze-only} & \textbf{0.280} & $[0.280, 0.280]$ & $+0.015$ & $< 0.001$ \\
Random & 0.265 & $[0.261, 0.270]$ & --- & --- \\
Gate+Random & 0.262 & $[0.258, 0.266]$ & $-0.003$ & $1.000$ \\
Naive fusion$^\ddagger$ & 0.256 & $[0.256, 0.256]$ & $-0.009$ & $1.000$ \\
Pupil-abs & 0.255 & $[0.255, 0.255]$ & $-0.011$ & $1.000$ \\
Dual & 0.253 & $[0.253, 0.253]$ & $-0.012$ & $1.000$ \\
\bottomrule
\end{tabular}
\end{table}

This asymmetry is a direct prediction of the quality--novelty decomposition. Activity recognition requires capturing \emph{transitions}, moments when behavior, objects, or the visual scene changes. Pupil dilation co-occurs with these novel moments, enriching the training set with transition frames that carry class-discriminative information. Scene recognition, by contrast, requires capturing \emph{place identity}~\cite{zhou2018places}, a spatial property fully determined by the visual content during stable fixation. Pupil novelty introduces variation orthogonal to scene identity. Specifically, gaze confidence dominates scene recognition (AULC $0.280$ vs.\ $0.253$ for dual), while pupil response dominates activity recognition ($\Delta$AULC $= +0.018$ for pupil ranking over gate+random). The fact that pupil helps temporal tasks and hurts spatial ones is the cleanest evidence that gaze and pupil carry genuinely different frame-level information.

\subsection{Temporal Alignment: Delayed vs.\ No-Delay}
\label{sec:temporal}

Given the biological latency of the pupil response (300--1500\,ms), should the pupil signal be aligned to the current frame (no-delay, centered window) or shifted forward to capture upcoming visual change (delayed, $+0.3$ to $+1.5$\,s)? We compare delayed and no-delay variants of the dual-criterion strategy across all six budgets for both tasks. Table~\ref{tab:temporal} summarizes the win counts.

\begin{table}[t]
\centering
\footnotesize
\caption{Temporal alignment. Delayed wins for activity; no-delay for scene.}
\label{tab:temporal}
\begin{tabular}{lccc}
\toprule
\textbf{Task} & \textbf{Delayed Wins} & \textbf{No-Delay Wins} & \textbf{Interpretation} \\
\midrule
Activity & 10 / 12 & 2 / 12 & Delay captures transition arousal \\
Scene & 4 / 12 & 8 / 12 & Scene is frame-local \\
\bottomrule
\end{tabular}
\end{table}

For activity recognition, the delayed pupil window wins 10 of 12 comparisons. The forward shift acts as a temporal smoother, capturing sustained arousal patterns associated with activity transitions rather than instantaneous fluctuations. For scene recognition, no-delay wins 8 of 12: scene identity is a frame-local property, and the current pupil state at the moment of fixation provides the most relevant curation signal.

This finding opens a design space for task-dependent temporal alignment of physiological signals. We tested two operating points; a continuous sweep over delay values is future work. The initial results suggest that optimal alignment follows the task's temporal structure: forward-looking for sequential tasks, frame-local for spatial ones.

%% file: sections/discussion.tex
\section{Discussion}
\label{sec:discussion}

\paragraph{What This Method Is and Is Not.}
Our goal is frame selection for sample-efficient learning, not cognitive state inference. Pupil dilation is not a label predictor: without gaze gating, raw pupil ranking performs poorly at the tight budgets (5--10\%) where on-device curation matters most, even though it becomes competitive at moderate budgets (25--50\%) once the quality floor is less critical. The signal's value lies in ranking frames \emph{within a quality-filtered pool}, where it surfaces moments of high feature change that random selection would miss. In this sense, the pupil channel is a curation signal, not a cognitive decoder. Unlike active learning~\cite{sener2018coreset} and submodular selection~\cite{gygli2015submod}, which require model forward passes over the entire stream, our curator operates at capture time using eye-tracking outputs available on recording hardware, before any model touches the data.

\paragraph{When It Works Best.}
Physiological curation is most effective for tasks where capturing transitions matters. For activity recognition, pupil-derived novelty highlights the state transitions that matter most for distinguishing activities. Accordingly, the dual curator achieves the highest AULC (0.223) and matches full-stream performance at just 10\% budget ($F1 = 0.228$ vs.\ $0.224$). Ungated pupil ranking excels at moderate budgets (25--50\%) but collapses at tight budgets where it selects noisy frames; the gaze gate eliminates this failure mode. For scene recognition, gaze-only selection already dominates; adding pupil ranking provides no benefit, consistent with the fact that scene identity is spatial and stable rather than transition-driven. The practical recommendation is to use the dual curator when the budget is tight and the task involves temporal variation, and to fall back to simpler gaze-only selection otherwise. When the frozen features themselves cannot separate two classes, such as phone vs.\ computer use ($F1 \approx 0.51$, near chance), no curation strategy helps. The bottleneck is the feature representation, not which frames are selected. We evaluate all main results with a linear probe on frozen DINOv2 features. To check robustness, we re-ran all strategies with MLP and gradient-boosted tree classifiers (supplementary). For activity recognition, naive fusion ranks near the bottom (5th of 6) under all three classifier families, confirming that fusing opposing signals into a single scalar destroys both contributions. Detailed strategy rankings are classifier-dependent (e.g., Gate+Random leads under MLP), which is expected when the probe evaluates frozen representations rather than learned features. For scene recognition, gaze-only ranks first under both Linear and MLP. Full classifier comparison results are provided in the supplementary material.

\paragraph{Limitations.}
All results are on VEDB; generalization to other egocentric datasets with synchronized eye tracking and pupil recording is untested. Existing large-scale egocentric benchmarks such as Ego4D~\cite{grauman2022ego4d} and EPIC-Kitchens~\cite{damen2022epic100} do not include pupil data, and datasets that do combine gaze with egocentric video~\cite{li2018egtea} lack pupil recordings, limiting direct cross-dataset validation at present. Using a frozen backbone limits absolute accuracy but ensures that performance differences reflect frame selection alone, not differences in model capacity. Pupil response varies with age, medication, and fatigue; cross-participant generalization requires further study. Eye tracking also raises privacy concerns~\cite{kroger2020gaze_privacy} that require explicit consideration in deployment.

%% file: sections/conclusion.tex
\section{Conclusion}
\label{sec:conclusion}

Eye-tracking signals already present on modern AR glasses can curate egocentric video streams without any vision model in the loop: our dual-criterion curator selects 10\% of frames that match full-stream classification accuracy for activity recognition. The key to making this work is composing gaze and pupil sequentially, as a quality gate followed by a novelty ranker, rather than fusing them into a single score. We demonstrate on the Visual Experience Dataset that a small fraction of curated frames matches full-stream classification performance, with the benefit being task-dependent: pupil ranking improves activity recognition while gaze-only selection suffices for scene recognition. Our work is a step toward leveraging built-in eye-tracking hardware on modern wearable devices for efficient, training-free data curation in egocentric vision and embodied AI.